\documentclass[conference]{IEEEtran}
\IEEEoverridecommandlockouts

\usepackage{cite}
\usepackage{amsmath,amssymb,amsfonts}
\usepackage{algorithmic}
\usepackage{graphicx}
\usepackage{textcomp}
\usepackage{xcolor}
\def\BibTeX{{\rm B\kern-.05em{\sc i\kern-.025em b}\kern-.08em
    T\kern-.1667em\lower.7ex\hbox{E}\kern-.125emX}}

\usepackage{hyperref}
\usepackage{todonotes}
\usepackage[utf8]{inputenc}
\usepackage{amsmath}
\usepackage{amsthm}
\usepackage[caption = false]{subfig}
\usepackage{xcolor}
\usepackage{colortbl}
\usepackage{amssymb}
\usepackage{dsfont}

\usepackage{pifont}
\usepackage{multirow}
\usepackage{booktabs}
\usepackage{diagbox}
\usepackage{microtype}
\usepackage{graphicx}
\usepackage{booktabs} 

\definecolor{gray1}{gray}{0.88}
\definecolor{gray2}{gray}{0.78}

\newtheorem{definition}{Definition}
\DeclareMathOperator*{\argmin}{arg\,min}

\newcommand\nnfootnote[1]{%
  \begin{NoHyper}
  \renewcommand\thefootnote{}\footnote{#1}%
  \addtocounter{footnote}{-1}%
  \end{NoHyper}
}

\makeatletter
\newcommand{\linebreakand}{%
  \end{@IEEEauthorhalign}
  \hfill\mbox{}\par
  \mbox{}\hfill\begin{@IEEEauthorhalign}
}
\makeatother

\begin{document}
\title{Imbalance in Regression Datasets}

\author{\IEEEauthorblockN{Daniel Kowatsch$^*$}
\IEEEauthorblockA{\textit{Fraunhofer AISEC} \\
Garching near Munich, Germany \\
daniel.kowatsch@aisec.fraunhofer.de}
\and
\IEEEauthorblockN{Nicolas M. Müller$^*$}
\IEEEauthorblockA{\textit{Fraunhofer AISEC} \\
Garching near Munich, Germany \\
nicolas.mueller@aisec.fraunhofer.de}
\and
\IEEEauthorblockN{Kilian Tscharke}
\IEEEauthorblockA{\textit{Fraunhofer AISEC} \\
Garching near Munich, Germany \\
kilian.tscharke@aisec.fraunhofer.de}
\linebreakand 
\IEEEauthorblockN{Philip Sperl}
\IEEEauthorblockA{\textit{Fraunhofer AISEC} \\
Garching near Munich, Germany \\
philip.sperl@aisec.fraunhofer.de}
\and
\IEEEauthorblockN{Konstantin Böttinger}
\IEEEauthorblockA{\textit{Fraunhofer AISEC} \\
Garching near Munich, Germany \\
konstantin.boettinger@aisec.fraunhofer.de}
}


\maketitle

\nnfootnote{$^*$ Equal contribution}









\begin{abstract}
For classification, the problem of class imbalance is well known and has been extensively studied. 
In this paper, we argue that imbalance in regression is an equally important problem which has so far been overlooked:

Due to under- and over-representations in a data set's target distribution, regressors  are prone to degenerate to naive models, systematically neglecting uncommon training data and over-representing targets seen often during training.
We analyse this problem theoretically and use resulting insights to develop a first definition of imbalance in regression, which we show to be a generalisation of the commonly employed imbalance measure in classification.
With this, we hope to turn the spotlight on the overlooked problem of imbalance in regression and to provide common ground for future research.
\end{abstract}

\begin{IEEEkeywords}
Machine Learning, Target Balance, Regression, Imbalance
\end{IEEEkeywords}

\section{Introduction}
The notion of data imbalance is an established concept in classification: A given classification data set is considered \emph{imbalanced} when, for two target classes $A$ and $B$, the frequency of occurrence is noticeably unequal.
Usually, we measure this imbalance as the quotient of the number of instances of the most common class (say class $A$) and the number of instances of the least common class (say class $B$).
Considering data imbalance is important, because otherwise models degenerate to trivial classifiers which primarily predict the majority class; thus, the prediction of less common classes suffers.
While this is common knowledge for classification, the problem has been largely ignored for regression, up to the point where there is not even a well defined notion of balance.
With this work, we highlight the problem of imbalance in regression.

In summary, our contribution is as follows:
\begin{itemize}
    \item We show that imbalance in regression data is a serious problem, which causes models to degenerate and only predict majority data.
    \item We review the literature and observe that this problem has, so far, been ignored by the scientific community.
    \item Thus, we take a first step towards solving this issue: we derive a mathematical measure of imbalance in regression which generalizes the established notion of imbalance in classification.
\end{itemize}

\section{Problem: Imbalance in Regression}\label{s:problem}
\begin{figure}
    \centering
    \includegraphics[width=0.9\linewidth]{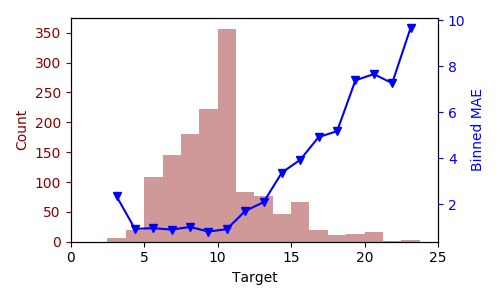}
    \caption{
    A histogram of the target variables of the  \href{https://archive.ics.uci.edu/ml/datasets/abalone}{abalone} training dataset (red bars).
    The blue line illustrates the mean absolute test error per histogram bin of a neural network trained on this data. 
    Note the heavy imbalance in the target variable.
    Because of this, rare values (i.e.~target $\geq 15$) are poorly predicted. The model degenerates to only predict targets in the interval $[5, 10]$.}
    \label{fig:intro01}
\end{figure}

In this section, we show the parallels between imbalance in classification and regression, the resulting degeneration to 'naive' models and the significance in real world data sets.

\subsection{Imbalance in Classification and Regression}\label{ss:intro_class_vs_reg}

\begin{figure*}[ht!]
\centering
    \begin{minipage}{.5\textwidth}
        \centering
        \begin{tabular}{l|cccc}
\toprule
\diagbox{M}{I} &           1  &           3  &           10 &           20 \\
\midrule
Accuracy &  .94$\pm$.03 &  .94$\pm$.01 &  .96$\pm$.01 & \cellcolor{blue!25} .97$\pm$.01 \\
TNR      &  .94$\pm$.02 &  .98$\pm$.01 &   .99$\pm$.00 &  \cellcolor{blue!25} .99$\pm$.01 \\
TPR      &  .93$\pm$.03 &  .85$\pm$.04 &   .68$\pm$.10 &  \cellcolor{red!25} .57$\pm$.10 \\
\bottomrule
\end{tabular}

        \caption{The impact of imbalance in classification data. \label{tab:class}}
    \end{minipage}%
    \begin{minipage}{0.5\textwidth}
        \centering
        \begin{tabular}{l|cccc}
\toprule
\diagbox{M}{I} &             1  &             3  &             10 &             20 \\
\midrule
MAE     &   .10$\pm$.02 &  .09$\pm$.01 &  .07$\pm$.02 & \cellcolor{blue!25} .05$\pm$.01 \\
$\text{MAE}_0$ &  .12$\pm$.05 &  .04$\pm$.01 &  .01$\pm$.01 & \cellcolor{blue!25}   .00$\pm$.00 \\
$\text{MAE}_1$ &  .09$\pm$.03 &  .23$\pm$.06 &  .67$\pm$.22 & \cellcolor{red!25}  .90$\pm$.19 \\
\bottomrule
\end{tabular}

        \caption{The impact of imbalance in regression data. \label{tab:reg}}
    \end{minipage}
    \caption*{
    This figure compares the effect of imbalance in classification and regression.
    The left table show an imbalanced classification dataset, where the degree of imbalance is given via $I$.
    As the imbalance increases, the model degenerates to always predict the majority class (see the low true-positive ratio TPR marked in red).
    However, looking only at the accuracy, this behavior cannot be inferred.
    Now consider the right table, which presents exactly the same scenario for a regression dataset.
    Observe that the the mean average error (MAE) for mode $1$ deteriorates dramatically, while the overall MAE does not.
    This highlights the fact that imbalance is just as severe in regression as it is in classifiction; albeit the fact that it has been largely ignored in the scientific community. Note: 
    Each entry corresponds to the mean and standard deviation over $10$ runs.}
\end{figure*}

In order to illustrate the parallels between imbalance in classification and regression, we consider a synthetic regression data set $(X, Y)$.
The distribution of the target variable $Y$ is bimodal, with one mode around $Y=0$ and one around $Y=1$.
For the classification task, we only focus on predicting the mode of the target variable.
To show the impact of data imbalance, we generate several synthetic data sets with different imbalance factors. The imbalance factor of a synthetic data set is given by the number of data points belonging to mode $0$ divided by the number of data points corresponding to mode $1$.
Note that this notion of imbalance for a regression data set is only reasonable due to the fact that the distribution of the target variable $Y$ is bimodal and target values outside the modes are considered irrelevant in this scenario. For general regression data sets, this notion is unsuited, and we will later suggest better suited approaches for quantifying imbalance in regression (see Section~\ref{s:measures}).

The experiments use imbalance factors (I) from $1$ to $20$ to generate synthetic data sets and train a neural network\footnote{
The neural network has three hidden layers, each with twenty neurons, and ReLU activations. The loss is either binary cross-entropy when used for classification, or MAE for regression.} for each.
We use a neural network, since it's a representative of a family of commonly used models.

First, we focus on the results for classification.
In this context we will refer to samples corresponding to mode $1$ as positives and to the remaining samples as negatives.
The averaged results and standard deviations for different imbalance factors are given in Table~\ref{tab:class}.


Regardless of the data imbalance, accuracy and True Negative Rate (TNR) are consistently high.
However, True Positive Rate (TPR) and F1 Score decrease as the data set becomes more imbalanced. 
For a high data-imbalance level (I $= 20$), the F1 Score deteriorates from $0.93$ to $0.68$.
Similarly, the TPR deteriorates from $0.93$ to $0.57$.
These results are consistent with common knowledge that class imbalance results in a reduction of the prediction quality for data samples belonging to minority mode $Y=1$. 

Now we consider the results for the closely related regression task (Table~\ref{tab:reg}).
As a general metric we use the Mean Absolute Error (MAE).
We observe that the MAE does not deteriorate even under considerable imbalance, just as the accuracy does not deteriorate for imbalanced classification.

To gain more insight into the regressor's behavior, we now consider the MAE of samples corresponding to mode $0$ ($\text{MAE}_0$) and to mode $1$ ($\text{MAE}_1$).
We then compare $\text{MAE}_0$ to the False Positive Rate in the classification problem, and $\text{MAE}_1$ to the True Positive Rate.
We observe the following:
Increasing the data imbalance results in a strong reduction of $\text{MAE}_0$ while $\text{MAE}_1$ increases. 
This indicates that as more data from mode $0$ is available, the model focuses on correctly predicting those samples
at the expense of correctly predicting samples of mode $1$.
Therefore, we can observe similar behaviour of the model for imbalance in classification and regression:
Compare this with the classifier's behavior in Table~\ref{tab:class}:
In both cases, the learner degenerates to primarily predicting values from mode $0$.
Values from mode $1$ are neglected, resulting in a naive prediction and ultimately rendering the model useless for predicting samples of mode $1$.
Also note that the MAE fails to record the model's degeneration (as does the accuracy).

In summary, we observe the same problems for regression and classification when dealing with imbalanced data and, thus, this problem should be considered in both tasks.


\subsection{Effects of Imbalance in Regression}\label{ss:intro_effects_imbalance}
\begin{figure}[ht]
    \centering
    \includegraphics[width=0.9\linewidth]{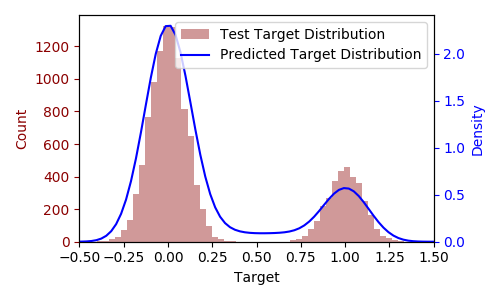}
    \includegraphics[width=0.9\linewidth]{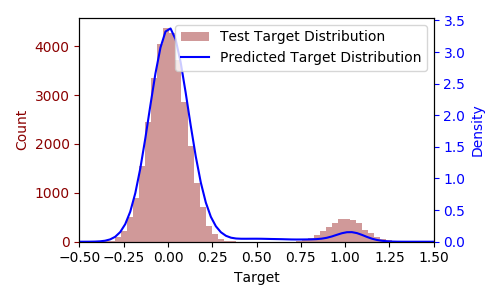}
    \includegraphics[width=0.9\linewidth]{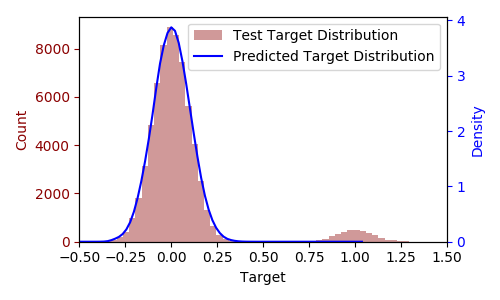}
    \caption{Histogram plots of three synthetic data sets, where the continuous target variable is bimodally distributed.
    For each figure, a  neural network regressor is trained on an equally imbalanced train set and evaluated on a test set (red bars).
    We plot the regressors predicted targets via KDE plot (blue line).
    The data set in the top figure has an imbalance factor of $3$, the middle one of $10$, and the bottom one of $20$.
    For higher degrees of imbalance, we observe the same phenomena as in classification:
    The regressors fails to capture the minority mode data and degenerates to a naive model.
    }
    \label{fig:reg_dists}
\end{figure}

In this subsection, we further analyse the running example from the last subsection. For this purpose consider Figure~\ref{fig:reg_dists}. 
It shows a distribution plot of the data set used in the regression example (red bars) for different degrees of data imbalance.
Train and test distribution are equally (im)balanced.
The blue line shows the regressors' predictions, drawn as a KDE plot.

Observe the top figure, which shows a data set with a moderate imbalance ratio of $3$. In this scenario, the regressor is able to fit both modes of the data distribution adequately, as indicated by the reasonable fit of the blue line to the red bars.
Now consider the image in the middle.
The imbalance is higher, i.e. there are ten data samples from mode $0$ for every sample of mode $1$. Observe that the predicted distribution fits worse to the data of mode $1$ which indicates a deterioration of the model.
Finally, consider the bottom image of Figure~\ref{fig:reg_dists}.
In this scenario the data has an imbalance factor of $20$, meaning data of mode $0$ is twenty times more frequent than data of mode $1$.
As we can see by the blue line representing the predicted data distribution, the model almost exclusively predicts values for mode $0$.
This reinforces our conclusion from Section~\ref{ss:intro_class_vs_reg}: Due to the imbalance present in the data set, the model loses the ability to predict data of mode $1$.

All of these data sets are generated from similar processes, the only difference is the imbalance factor.
In order to eliminate information loss as an influencing factor, we modify the imbalance factor by increasing the number of data samples belonging to mode $0$ while the number of data points for mode $1$ stays constant.
For a classification task, we would now conclude: 
The model has not learned adequately due to class imbalance. 
We argue that a similar verdict should be applied to the regression learner.
However, thus far, this issue has been ignored. There is no established analogue to Balanced accuracy or F1 Score for regression.
There is not even a metric to quantify imbalance in regression or a proper definition of balance, making it hard for data scientists to even describe the problem / phenomenon, or to determine before training if counter measures might be needed.

Before we proceed to suggest such an imbalance metric for regression, we require a better understanding of the problem at hand.
For this purpose, we analyse the training loss theoretically.
So, let $\mathcal{X}$ be the feature domain, $\mathcal{Y}$ the target domain and $X$, $Y$ be random variables with joint density $p_{X,Y}$ describing the problem at hand.

Further, let $f:\mathcal{X}\to\mathcal{Y}$ be a model describing some point estimate of $p_{Y|X}$ and $\mathcal{L}:\mathcal{Y}\times\mathcal{Y}\to\mathbb{R}$ be a loss function, e.g. the squared error.
Then the loss of the model can be considered as an expected loss given by
\begin{align}
    \mathbb{E}[\mathcal{L}] &= \int_{\mathcal{Y}}\int_{\mathcal{X}} \mathcal{L}(y, f(x))p_{X,Y}(x,y)dxdy\\
    &=\int_{\mathcal{Y}}\int_{\mathcal{X}} \mathcal{L}(y, f(x))p_{X|Y}(x|y)dx\, p_Y(y)dy\\
    &= \int_{\mathcal{Y}} \hat{\mathcal{L}}(y)p_Y(y)dy,
    \label{eq:expected_loss_simple}
\end{align}
where $\hat{\mathcal{L}}(y)= \int_\mathcal{X}\mathcal{L}(y, f(x))p_{X|Y}(x|y)dx$.
For a given point in time during optimisation, we can consider the model $f$ to be fixed.
Furthermore, in practical application the amount of data for an estimation of $p_{X|Y}$ is limited, so we primarily focus on the density $p_Y$, which is easier to estimate.

Based on Eq.~\ref{eq:expected_loss_simple}, we see that intervals with higher density have stronger influence on the expected loss.
Thus, model training algorithms like gradient descent methods will optimise the model towards intervals with higher density.

\cite{Branco:2016:SPM:2966278.2907070} have argued that imbalance is only a problem if poorly represented intervals are considered relevant. 
This can be observed in our example (Figure~\ref{fig:reg_dists}), where we are only interested in predicting data from mode 0 and 1.
Hence, a lack of training data with targets in $[1.5, \infty)$ is not problematic, and should not contribute towards the imbalance factor.
This foreshadows results from Section~\ref{s:balance_def}, where we argue that imbalance should depend on a notion of relevance with regards to the target domain.

In summary, we conclude:
\begin{itemize}
    \itemsep0em
    \item Intervals in the target domain with higher density have a stronger influence on the expected loss and, thus, on model optimisation.
    \item Quantifying balance and imbalance requires some notion of relevance.
\end{itemize}

\subsection{Imbalance in Real World Data}

In this section, we present a real-world example of imbalance in regression data.
This indicates that imbalance in regression is not only a theoretical phenomenon, but that it has a tangible impact.
We thusly argue that the ML community should start to take imbalance in regression into consideration.

Consider Figure~\ref{fig:intro01}, which shows the target variable distribution of the real-world $abalone$ data set~\cite{waugh1995extending, Dua:2019}. 
As can be seen, $target$ values of around 10 are represented extensively. 
For higher $target$-values, the number of data points drops, and between 15 and 25, there is only very little data. 
We argue that a regressor trained 'naively' on this data would develop a bias towards the common targets $\in [5, 15]$, and have a tendency to predict values outside this range poorly.
To evaluate this, we split the data into train/test sets, and train  a neural network as in Section~\ref{ss:intro_class_vs_reg}.
Finally, we evaluate the MAE on the test set \emph{per bin}.
This means that, for $B=20$ bins as shown in the figure, we calculate $20$ MAE scores, based on the true target values.
For example, given some $y_k = 17$ and the regressor's corresponding prediction $pred_k = 11$, an Absolute Error of $17-11 = 6$ will be attributed to bin $17$.
Formally: For each bin defined by an interval $I_j=[b_j, b_{j+1}]$ the corresponding MAE score is given by
\begin{equation}
    MAE_j = \frac{1}{S_j}\sum_{i=1}^N \mathds{1}_{I_j}(y_i)|y_i - pred_i|, 
\end{equation}
where $N$ is the number of samples in the data set, $y_i$ a true target value, $pred_i$ a predicted target value and $\mathds{1}_{I_j}$ is the indicator function on interval $I_j$.
$S_j = \sum_{i=1}^N \mathds{1}_{I_j}(y_i)$ counts the number of samples in bin $j$.

We do this to measure if the regressor predicts common values better than rare values. As shown in Section~\ref{ss:intro_class_vs_reg}, in an imbalanced scenario, the standard MAE is a limited indicator as to whether or not the regressor has learned to accurately predict the continuous targets. This is why we report the binned MAE.

We plot the resulting MAE scores per bin as a blue line.
We observe that for common values (i.e. $T=10$), there is low error.
For uncommon values however ($T > 20$), there is high error.
We see that for $T>10$, the error increases almost linearly.
The regressor has learned to always predict the mode of the distribution. It has not learned to predict uncommon targets.
Analogously to the synthetic data set presented in Section~\ref{ss:intro_effects_imbalance}, we argue that this behavior may not be desirable, and that we should develop the tools to quantify and address it.

\section{Definition of Balance}
\label{s:balance_def}

Based on the insights from Section~\ref{s:problem}, we define a notion of balance in this section.
To get started, we consider balance in classification. 

\subsection{Definition of Target Balance for Classification}
\label{ss:def_balance_class}
A classification problem with finite class domain $\mathcal{Y}$ is considered balanced if the probability mass associated with each class is equal.
Formally:
\begin{definition}
    \label{def:class_balance}
    Let $\mathcal{Y}$ with $|\mathcal{Y}| < \infty$ be the class domain of a classification problem given by a joint probability mass function $p_{X,Y}$, $Y$ a random variable with probability mass function $p_Y$, which describes the marginal probability distribution of the problem on the class domain.
    Then the classification problem is considered balanced if
    \begin{equation}
        \forall y, y' \in \mathcal{Y}:p_Y(y)=p_Y(y').
    \end{equation}
\end{definition}

The next goal is to define the notion of 'balance' for continuous target variables, i.e. for regression problems.
The simplest approach would be to consider a continuous target domain $\mathcal{Y}$ and a corresponding random variable $Y$ instead of a discrete set of classes and substitute the probability mass function for the density function of $Y$.
However, this approach is only suited for a subset of all probability measures, since there exist probability measures, for which no probability density function exists, e.g. the Cantor distribution.
So, we will utilise probability measures instead.

There is one more issue to observe: Namely that we may care more about certain intervals than others. 
We will investigate how to deal with this in the following Section.

\subsection{Target Relevance in Regression}
\label{ss:relevance_measure}
In Section~\ref{ss:intro_effects_imbalance}, we argued that the notion of balance in a continuous domain depends on some notion of relevance.
For example, in Figure~\ref{fig:reg_dists}, we were not interested in predicting targets in the interval $[1.5, \infty)$, and, thus, we did not mind the lack of corresponding training samples.
On the other hand, however, we did care about targets in the proximity of $Y=1$, so the lack of corresponding training samples was an issue.
Therefore, in order to define imbalance in regression and tackle the issue mentioned in Section~\ref{ss:def_balance_class}, we need to define a notion of relevance.

We define relevance using measures.
But before we formalise relevance measures, we want to point out that
the choice of relevance measure depends on the task at hand and the goal to be achieved.
This also means there is no universal relevance measure which is always reasonable to use, but there might be some commonly useful ones.

Let $\mathcal{P}(\mathcal{Y})$ be the power set of $Y$, and let 
$\mathcal{A}\subseteq\mathcal{P}(\mathcal{Y})$ be a $\sigma$-algebra on $\mathcal{Y}$, then a relevance measure is a function $\mu$ of the form 
\begin{equation}
    \mu:\mathcal{A}\to\mathbb{R}_0^+ \cup \{\infty\}.
\end{equation}
Intuitively, we map a subset of the target domain $Y$ to some relevance score $\geq 0$.
Since we use probability measures for the generalisation of balance, it is convenient to use functions that map sets to relevance scores.
We now proceed to argue for the properties of measures.

\subsubsection{Non-negativity}
First, we require that a relevance measure is non-negative, i.e.
\begin{equation}
    \forall S\in\mathcal{A}: \mu(S)\geq 0.
    \label{eq:nonneg}
\end{equation}
Thus, for a set $S\in\mathcal{A}:\mu(S) = 0$ implies that $S$ is irrelevant with respect to the relevance measure.

\subsubsection{Null Empty Set}
Second, the empty set $\emptyset$ should be considered as irrelevant since it contains no (relevant) elements, so
\begin{equation}
    \mu(\emptyset) = 0.
    \label{eq:emptyzero}
\end{equation}

\subsubsection{Countable Additivity}
Third, we require that the relevance of any countable union of disjoint sets equals the sum of the relevance of the individual sets.
Formally, this means that for any sequence of disjoint sets $\{A_i\}_{i\in\mathbb{N}}$
\begin{equation}
    \mu(\cup_{i\in\mathbb{N}}A_i) = \sum_{i\in\mathbb{N}}\mu(A_i)
    \label{eq:sigma_add}
\end{equation}
holds.
This is motivated by the fact that the influence of disjoint intervals on the optimisation problem given in Eq.~\ref{eq:expected_loss_simple} sums up and, thus, for simplicity the relevance of disjoint intervals should sum up, too.

\subsection{Generalised Definition of Target Balance}
Flowing from the notion of balance in classification, we propose the following generalised definition of balance, which applies to both categorical and continuous targets.

\begin{definition}
    \label{def:generalized_balance}
    Let $\mathcal{Y}$ be the target domain of some problem given by a joint probability measure $P_{X, Y}$, $Y$ a random variable with marginal probability measure $P_Y$ describing the problem on $\mathcal{Y}$.
    Further let $\mathcal{A}$ be a $\sigma$-algebra on $\mathcal{Y}$ and $\mu:\mathcal{A}\to\mathbb{R}_0^+\cup\{\infty\}$ be a relevance measure on $\mathcal{Y}$.
    Then the target domain is $\mu$-balanced if
    \begin{equation}
        \label{eq:generalized_balance}
        \forall S, S' \in \mathcal{A}: \mu(S) \leq \mu(S') \implies P_Y(S) \leq P_Y(S'),
    \end{equation}
\end{definition}
where $P_Y(S) = P_Y(y \in S)$ denotes the probability that the random variable $Y$ takes on any value in the set $S \in\mathcal{A}$.
Thus, a target domain is considered $\mu$-balanced if higher relevance according to $\mu$ implies higher probability and, thus, higher influence on the expected loss and the optimisation process.

\subsection{Special Cases of Balance}
Using the generalised definition of $\mu$-balance we now consider some special cases in order to illustrate some of the implications of our definition.
If not stated otherwise in an example, let $\mathcal{Y}\subseteq\mathbb{R}$ be a continuous target domain, $Y$ a random variable with probability measure $P_Y$ describing the problem at hand on $\mathcal{Y}$ and $\mathcal{A}$ a suitable $\sigma$-algebra.

First, we show that the definition of balance for classification given in Definition~\ref{def:class_balance} can be considered as a special case of our generalised definition of balance formalised in Definition~\ref{def:generalized_balance}.
For this purpose consider a classification task with finite class domain $\mathcal{Y}$, i.e. $|\mathcal{Y}| < \infty$, and the count-measure $\mu$ which assigns each set the number of elements in it.
Using the count-measure $\mu$ as relevance measure Eq.~\ref{eq:generalized_balance} simplifies to
\begin{equation}
     \forall S, S' \in \mathcal{A}: |S| \leq |S'| \implies P_Y(S) \leq P_Y(S').
\end{equation}
So, considering single-element sets yields Definition~\ref{def:class_balance}.
Since a discrete probability measure is defined by the probability masses of each element, Definition~\ref{def:class_balance} also implies Definition~\ref{def:generalized_balance} for $\mu$-balance in this special case.
Therefore, both definitions are equivalent and the commonly used definition of balance in classification can be considered as a special case of $\mu$-balance.

Second, we present an example which shows the importance of choosing an appropriate relevance measure.
Let us consider a regression problem on some arbitrary target domain.
In case there are no preferences for values in certain intervals, it is reasonable that the relevance of an interval only depends on its length.
To capture this, the relevance measure will be the Lebesgue measure $\lambda$, which assigns each interval $(a,b)$ its natural length $b-a$.

In case $\mathcal{Y}$ is bounded this results in $\lambda$-balance if $Y$ is uniformly distributed on $\mathcal{Y}$.
This can be seen by considering intervals of equal length in Eq.~\ref{eq:generalized_balance}.
Note, that if $\mathcal{Y}$ is not bounded, there is not necessarily a probability measure $P_Y$ which satisfies $\lambda$-balance.
To see this, consider $\mathcal{Y}=\mathbb{R}$ and the set of intervals $\mathcal{I}=\{(n, n+1)| n\in\mathbb{Z}\}$.
Each interval $I\in\mathcal{I}$ has length $1$ and therefore its Lebesgue measure is also $1$.
So, in order for $P_Y$ to satisfy Definition~\ref{def:generalized_balance} each interval $I \in \mathcal{I}$ also has to have equal probability under $P_Y$.
But since $\cup_{I \in \mathcal{I}}I = \mathbb{R}$ any probability $P_Y(I)>0$ would result in
\begin{align}
    P_Y(\mathbb{R}) &= P_Y(\cup_{I\in\mathcal{I}}I)\\
    &= \sum_{I\in\mathcal{I}}P_Y(I) = \infty
\end{align}
contradicting that $P_Y$ is a probability measure.
In case $\forall I \in \mathcal{I}: P_Y(I) = 0$, then the probability 
\begin{align}
    P_Y(\mathbb{R}) &=P_Y(\cup_{I\in\mathcal{I}}I)\\
    &= \sum_{I\in\mathcal{I}}P_Y(I) = 0
\end{align} also contradicts that $P_Y$ is a probability measure on $\mathbb{R}$.
This example shows that whether or not $\mu$-balance can be achieved depends on the used relevance measure $\mu$ and the target domain $\mathcal{Y}$.

Third, we consider a trivial case where we use $\mu = P_Y$ as a relevance measure.
In this case Eq.~\ref{eq:generalized_balance} from the definition of $\mu$-balance simplifies to
\begin{equation}
    P_Y(S) \leq P_Y(S') \implies P_Y(S) \leq P_Y(S'),
    \label{eq:dist_balance}
\end{equation}
where $S, S' \in \mathcal{A}$ arbitrary.
Since Eq.~\ref{eq:dist_balance} always holds, any problem  is always balanced with regards to its probability measure.

\section{Quantification of Imbalance in Regression}\label{s:measures}

In Section~\ref{s:balance_def} we have introduced a definition for $\mu$-balance of the target variable, but in a real-world environment, we cannot directly apply Definition~\ref{def:generalized_balance}.
This is because we usually have neither the true data distribution function$F_{X,Y}$ nor the marginal distributions $F_X$ and $F_Y$, but only a data set $\mathcal{D}=(X_\mathcal{D}, Y_\mathcal{D})$.
Thus, we need to  approximate $F_Y$, using for example the empirical distribution function.
In the following chapter, we list requirements for imbalance quantification, and then proceed to suggest possible suitable candidates.

\subsection{Requirements for Imbalance Quantification}\label{ss:req_for_imb}

Before we discuss approaches to the quantification of imbalance in regression,
we present some theoretical and practical requirements.

In general, an approach to quantifying imbalance with respect to some relevance measure $\mu$
should map a relevance measure $\mu$ and a probability measure $P_Y$ to an imbalance score $d\in \mathbb{R}\cup \{\pm\infty\}$.
Formally:
\begin{equation}
    d:\mathcal{M}\times\mathcal{P}\to\bar{\mathbb{R}},
\end{equation}
where $\mathcal{M}$ is the set of all measures on $\mathcal{Y}$ and $\mathcal{P}$ is the set of probability measures on $\mathcal{Y}$.
We will refer to such a function $d$ as imbalance function.
Next, we argue for some additional requirements.

\subsubsection{Minimal Imbalance}
First, an adequate imbalance metric should only assume its global minimum if and only if a probability measure $\nu$ is $\mu$-balanced.
Formally: Let $\mu$ be a relevance measure and $\nu$ a probability measure. Then
\begin{align}
    \label{eq:balance_implies_min}
    \nu \textit{ is } \mu{\text{-}balanced}
    \iff 
    \mu, \nu \in \argmin_{\bar{\mu}\in\mathcal{M}, \bar{\nu}\in\mathcal{P}}d(\bar{\mu}, \bar{\nu}).
\end{align}

\subsubsection{Comparability}
Second, we require that the infimum and supremum of all imbalance functions $d_\mu = d(\mu, \cdot)$ is equivalent.
Formally:
\begin{align}
     \label{eq:identical_inf}
    \forall \mu, \bar{\mu} \in \mathcal{M}: \inf_{\nu \in \mathcal{P}}d(\mu, \nu) = \inf_{\nu \in \mathcal{P}}d(\bar{\mu}, \nu),\\
     \label{eq:identical_sup}
    \forall \mu, \bar{\mu} \in \mathcal{M}: \sup_{\nu \in \mathcal{P}}d(\mu, \nu) = \sup_{\nu \in \mathcal{P}}d(\bar{\mu}, \nu).
\end{align}
This allows for comparable value ranges regardless of $\mu$ and, thus, eases comparability.

\subsection{Examples for Imbalance Metrics}
In this section, we provide some suitable candidates for imbalance metrics, subject to the condition that the relevance measure $\mu$ is finite, i.e. $\mu(\mathcal{Y}) < \infty$.

Without loss of generality we will assume $\mu$ to be a probability measure.
This is possible due to the fact that normalisation of finite measures results in probability measures.
This allows us to consider established metrics for probability measures in the context of imbalance quantification.

In the following, let $\mathcal{Y}=\mathbb{R}$ be the target domain and $Y$ a random variable with probability measure $P_Y$ and probability distribution function $F_Y$, describing the marginal distribution of a given problem on the target domain $\mathcal{Y}$.
Further, let $F_\mu$ be the probability distribution function associated with the relevance measure $\mu$.

\subsubsection{Kolmogorov Metric}
The first candidate for quantifying imbalance is the Kolmogorov metric~\cite{Gibbs2002}, defined as
\begin{equation}
    d_{kol}(\mu, P_Y) = \sup_{x\in\mathbb{R}}|F_\mu(x)-F_Y(x)| \in [0, 1].
\end{equation}
To make use of the Kolmogorov metric, we need to approximate the probability distribution functions $F_\mu$ and $F_Y$, given the data $\mathcal{D}$.
$F_Y$ can be approximated by the empirical distribution function which converges almost surely towards $F_Y$ uniformly, given the samples are i.i.d.
This also allows for efficient approximation of the supremum.

We now proceed to show that the requirements presented in Section~\ref{ss:req_for_imb} hold for the Kolmogorov metric $d_{kol}$.
First, note that $d_{kol}$ of two probability measures is $0$ if and only if both distributions are equivalent.
Thus, the Kolmogorov metric satisfies Eq.~\ref{eq:balance_implies_min} if limited to probability measures.
Eq.~\ref{eq:identical_inf} is also satisfied, since for all finite relevance measures $\mu$ there exists a $\mu$-balanced probability distribution function, e.g. the one defined by normalising $\mu$.
Next, we show that Eq.~\ref{eq:identical_sup} holds with $\sup_{\nu\in\mathcal{P}} d_{kol}(\mu, \nu) = 1$.
Since $F_\mu$ is a probability distribution function,
\begin{equation}
    \lim_{x \to \infty} F_\mu(x) = 1
\end{equation}
holds.
Thus, for each $n\in\mathbb{N}$ there exists $x_n\in\mathbb{R}$ such that
\begin{equation}
    F_\mu(x_n)\geq1-\frac{1}{n}
\end{equation}
Next, let $\nu_n$ be some probability measure with $F_{\nu_n}(x_n) = 0$, e.g. with uniform distribution on $[x_n, x_n{+}1]$.
It follows that
\begin{equation}
    d_{kol}(\mu, \nu_n) \geq 1-\frac{1}{n}
\end{equation}
holds.
Therefore, there exists a sequence of probability measures $\nu_n$ with
\begin{equation}
    \lim_{n \to \infty}d_{kol}(\mu, \nu_n) = 1 \leq \sup_{\nu \in \mathcal{P}}d_{kol}(\mu, \nu)
\end{equation}
and since the Kolmogorov metric is upper bounded by $1$, it follows that Eq.~\ref{eq:identical_sup} holds.

\subsubsection{Wasserstein Metric}\label{ss:w2u}
The second candidate that offers itself for quantifying imbalance in regression is the Wasserstein metric~\cite{Gibbs2002}.
For real valued target domains it is defined as
\begin{equation}
    d_{wst}(\mu, P_Y) = \int_{-\infty}^\infty|F_\mu(x)-F_Y(x)|dx.
\end{equation}

Similar to the Kolmogorov metric, the Wasserstein metric requires the calculation or approximation of the distribution functions $F_\mu$ and $F_Y$.
In practical application $F_Y$ can be approximated using samples.
Thus, the integral can also be approximated numerically.

That Eq.~\ref{eq:balance_implies_min} and Eq.~\ref{eq:identical_inf} hold for the Wasserstein metric follows from these equations holding for the Kolmogorov metric.

It remains to show that for the Wasserstein metric Eq.~\ref{eq:identical_sup} holds with $\sup_{\nu \in \mathcal{P}}d_{wst}(\mu, \nu) = \infty$.
To show this, consider any finite relevance measure $\mu$ and its probability distribution function $F_\mu$.
Since
\begin{equation}
    \lim_{x\to\infty}F_\mu(x) = 1
\end{equation}
there exists some $x_0$ with
\begin{equation}
    F_\mu(x_0)\geq 0.9\text{.}
\end{equation}
For $\epsilon > 0$, let $\nu_\epsilon$ be a probability measure with a distribution function $F_\epsilon$ for which $F_\epsilon(x_0+\epsilon) = 0$ holds, e.g. the uniform distribution on $[x_0{+}\epsilon, x_0{+}\epsilon{+}1]$.
Then it follows that
\begin{align}
    0.9\epsilon &\leq \int_{x_0}^{x_0+\epsilon}|F_\mu(x) - F_\epsilon(x)|dx\\
    &\leq \int_{-\infty}^\infty|F_\mu(x) - F_\epsilon(x)|dx \\
    &= d_{wst}(\mu, \nu_\epsilon).
\end{align}
The limit $\epsilon\to\infty$ then results in the Wasserstein metric being unbounded for fixed $\mu$.
From this follows that Eq.~\ref{eq:identical_sup} holds with $\sup_{\nu\in\mathcal{P}}d_{wst}(\mu, \nu) = \infty$.

\subsection{Empirical Evaluation of Imbalance Metrics}

The main goal of using imbalance metrics is to estimate whether a model will be negatively impacted by a $\mu$-imbalanced data set.
For this purpose, we calculate the Pearson correlation coefficient between the imbalance scores for a data set given a set of imbalance measures $\mu_i$ and an performance evaluation metric for a model.

In our experiments, the model is a neural network and the architecture is identical to the experiments in Section~\ref{s:problem}.
We use four evaluation metrics to estimate the performance of a model given an imbalanced data set.
The first one is a weighted variant of the Mean Absolute Error, where the weights for a sample $y_i$ is given by $\frac{p_\mu(y_i)}{p_Y(y-i)}$. Here, $p_\mu$ is the density of the relevance measure $\mu$ and $p_Y$ is the density of the target variable of the data set, which is estimated using histograms with $20$ bins.
The second and third one are Precision and Recall for regression~\cite{torgo2009precision}.
We use an error threshold of $10$, a relevance threshold of $0.5$, and $k=1$.
For the relevance function we use the density of the relevance measure scaled to the value range $[0, 1]$.
The last one is the F score for regression~\cite{Branco2014ResamplingAF} with $\beta = 1$, which mirrors the F1 score for imbalanced classification.

For each data set we estimate the data set imbalance and the performance of a model using $20$ different normal distributions as relevance measures.
Their mean and standard deviation are equidistant points on the straight from the data set's mean and standard deviation and an endpoint, which is arbitrarily selected in less probable ranges of the target value range.

We run the experiment using three different data sets.
The first one is Abalone~\cite{waugh1995extending, Dua:2019} for which we use a mean of $18$ and a standard deviation of $5$.
The second one is a medical data set on Warfarin dosages~\cite{PharmGKB_2019}.
As endpoint of the series of relevance measures, we use a normal distribution with mean $80$ and standard deviation $10$.
The third data set is another medical data set used for early stage Parkinson detection~\cite{parkinson}.
Here, we work with an endpoint with mean $50$ and standard deviation $7$.

The aforementioned experiment is run 10 times. The Pearson correlation coefficients for all pairs imbalance metric/evaluation metric is given in Table~\ref{tab:imb_corr}.
Based on the results, we suggest using the Wasserstein distance as an imbalance metric, since it has very strong correlations for the weighted MAE, Recall and F1 score and a strong correlation for precision.

\begin{table}[b]
    \centering
    \caption{Pearson correlation between imbalance metrics and balanced model performance metrics.}
    \begin{tabular}{lrrrr}
\toprule
{} &     Weighted MAE &  Precision &    Recall &        F1 \\
\midrule
Kolmogorov  &  0.627576 &  -0.330374 & -0.582538 & -0.478573 \\
Wasserstein &  0.891372 &  -0.756323 & -0.883840 & -0.859176 \\
\bottomrule
\end{tabular}

    \label{tab:imb_corr}
\end{table}

\section{Related Work}
\label{s:related_work}
Since the problem of imbalance in regression has so far been largely overlooked, there exists very little work on the topic.
Previous work has mainly been done in the domain of relevance functions~\cite{Krawczyk2016}, which are primarily used for generating cost functions.
These relevance functions should not be confused with the relevance measures we introduce.
They assign relevance to each $y\in\mathcal{Y}$, not subsets $S\subseteq\mathcal{Y}$ like relevance measures do.
As a result they are suited for relevance-aware loss functions, but not for a definition of balance.
Additionally, sampling approaches have been explored, which try to create a more balanced data set with respect to groups of samples.
\cite{Branco:2016:SPM:2966278.2907070} provide an overview of methods dealing with imbalanced data sets, both for classification as well as regression tasks. 

\subsection{Relevance Functions}\label{ss:relevancy_fncts}
A common way to deal with imbalanced regression is by introducing a relevance function. 
This is also known as utility-based regression. 
Here, the loss is weighted by some relevance function $u$.
This relevance function is mostly crafted by hand, but may also be inferred from the target variable's density \cite{torgo2009precision}.
This approach is often used in conjunction with dividing the target range into two subsets:
One, where correct predictions are absolutely critical, and one where they are not.
A motivating example is stock trading, where large changes in the stock value are of high interest. 
However, predicting minor changes correctly is irrelevant, since these do not indicate to buy or sell a given stock value due to the associated financial overhead.

Building upon utility based regression, Torgo \& Riberio~\cite{torgo2009precision} present \emph{Precision and Recall for Regression}. They borrow the idea from binary classification, and use a relevance function and a threshold to divide the target range into relevant and irrelevant subdomains.

Jos{\'{e}} Hern{\'{a}}ndez-Orallo~\cite{Hernandez-Orallo2013} presents \emph{ROC curves for regression}, which employs the notion of a $y$-shift to balance out over- and underestimation in a regression model.
This allows for adjusting the model's over- and underestimation to an asymmetric loss.

Yang et al. \cite{yang2021smoothing} propose distribution smoothing for labels and features by using a symmetric kernel. Label distribution smoothing estimates the effective label density distribution which can be used to create a relevance function for the loss. Feature distribution smoothing on the other hand adjusts the statistics of the features by smoothing the feature covariance and mean.

\subsection{Sampling Methods}
Another approach when dealing with imbalanced regression data is to use sampling methods that either under- or over-sample, or synthesise new data.
These methods are also called \emph{preprocessing methods}, and have the advantage of not having to modify the regression algorithm itself.

In this context, Torgo et al.~\cite{10.1007/978-3-642-40669-0_33} adapt the well-known SMOTE algorithm~\cite{Chawla:2002:SSM:1622407.1622416} for regression.
Inspired by binary classification, they use a relevance function and a threshold to define which samples are \emph{rare} and which are \emph{normal}. 
Smote-R then synthesises new \emph{rare} samples using an interpolation strategy while downsampling \emph{normal} samples.

Branco et al.~\cite{Branco2017} present another sampling approach called SMOGN that addresses shortcomings Smote-R and creates a more diverse set of synthetic data. 
However, the underlying approach to deal with imbalance in regression by synthesising new \emph{rare} samples and downsampling \emph{normal} ones remains the same. 

Wang and Wang \cite{wang2023vir} introduce `Variational Imbalanced Regression', an autoencoder that samples similar data points to calculate the latent representation's variational distribution. In addition, the decoder performs probabilistic reweighting on the imbalanced data.

\section{Conclusion and Future Work}
\label{s:conclusion}
In this paper, we have shown that imbalance in regression can cause problems comparable to imbalance in classification and, thus, should be considered when designing models for regression.
Using the insights from a theoretical analysis of the problem we generalise the notion of balance to continuous target domains by utilising relevance measures.
Additionally, we suggest first candidates for quantifying imbalance.

In order to fully establish imbalance metrics, additional work is required.
Further, our insights might be used to find models or optimisation processes which are more robust against target imbalance and to develop counter measures to negate the effects of imbalance in regression.

\bibliography{regressionBib}
\bibliographystyle{IEEEtran}

\end{document}